\documentclass{svproc}
\usepackage{graphicx}
\usepackage{url}

\usepackage{algorithmic}
\usepackage[ruled, vlined, linesnumbered]{algorithm2e}

\usepackage{tikz}
\def\checkmark{\tikz\fill[scale=0.3](0,.35) -- (.25,0) -- (1,.7) -- (.25,.15) -- cycle;}

\usepackage{bm}

\usepackage{graphics}
\usepackage{trimclip} 

\usepackage{enumitem}

\usepackage{xr}

\usepackage{hyperref}
\usepackage{url}
\usepackage{amsmath}
\usepackage{amssymb}

\usepackage{amsfonts}
\usepackage{dsfont}
\usepackage{upgreek}
\usepackage{bbm}			

\usepackage{graphicx}
\usepackage{epstopdf}
\graphicspath{ {./images/} } 
\usepackage{tikz}
\usepackage{float}

\usepackage{booktabs} 

\usepackage[labelfont={small,bf}, textfont={small}]{caption}
\usepackage{subcaption}

\usepackage{xspace}

\usepackage[scaled=.78]{beramono}

\makeatletter
\g@addto@macro\bfseries{\boldmath}
\makeatother

\newlength{\phaserulewidth}
\newcommand{\setphaserulewidth}{\setlength{\phaserulewidth}}

\makeatother
\setphaserulewidth{.3pt}

\newcommand{\overbar}[1] {\mkern 1.5mu\overline{\mkern-3.5mu#1\mkern-1.0mu}\mkern 1.5mu}

\newcommand{\Sec}[1]		{Sec.\,\ref{#1}}
\newcommand{\Fig}[1]		{Fig.\,\ref{#1}}

\newcommand{\Eq}[1]			{Eq.\,\ref{#1}}
\newcommand{\Tab}[1]		{Tab.\,\ref{#1}}
\newcommand{\Alg}[1]		{Alg.\,\ref{#1}}

\newcommand{\vs}   			{vs.\@\xspace}
\newcommand{\ie}   			{i.e.\@\xspace}
\newcommand{\eg}   			{e.g.\@\xspace}

\newcommand{\mydots} 	{...}

\newcommand{\dset}[1]  {\texttt{#1}} 

\newcommand{\inlinetitle}[2]  {\vspace{4pt}\noindent\textbf{\emph{#1}{#2}}}

\DeclareMathOperator*{\argmax}{\arg\!\max}

\usepackage[normalem]{ulem} 

\newcommand{\mySqBullet}		{\raisebox{0.25em}{{\scriptsize$_\blacksquare$}}}

\newcounter{marginNoteCounter}

\usepackage[labelfont={footnotesize,bf}, textfont={footnotesize}]{caption}

\newcommand{\agreement}{\mathcal{A}}
\newcommand{\performance}{\mathcal{M}}

\usepackage{multirow}

\begin{document}
\mainmatter              
\title{To tree or not to tree? Assessing the impact of smoothing the decision boundaries}
\titlerunning{To tree or not to tree?}  
%
\author{Anthea Mérida \and Argyris Kalogeratos \and Mathilde Mougeot}
\authorrunning{Mérida et al.} 
%
\tocauthor{Anthea Mérida, Argyris Kalogeratos, Mathilde Mougeot}
\institute{\textit{Centre Borelli},
\textit{ENS Paris-Saclay,} \textit{Universit\'e Paris-Saclay}, France \\
\email{\smaller name.surname@ens-paris-saclay.fr\\}}

\maketitle              

\begin{abstract}

When analyzing a dataset, it can be useful to assess how smooth the decision boundaries need to be for a model to better fit the data. This paper addresses this question by proposing the quantification of how much should the `rigid' decision boundaries, produced by an algorithm that naturally finds such solutions, be relaxed to obtain a performance improvement. The approach we propose starts with the rigid decision boundaries of a seed Decision Tree (seed DT), which is used to initialize a Neural DT (NDT). The initial boundaries are challenged by relaxing them progressively through training the NDT. During this process, we measure the NDT's performance and decision agreement to its seed DT. We show how these two measures can help the user in figuring out how expressive his model should be, before exploring it further via model selection. The validity of our approach is demonstrated with experiments on simulated and benchmark datasets.

\end{abstract}
\section{Introduction}\label{sec:intro}

\begin{figure}[t]\small
\centering
\includegraphics[width=0.5\columnwidth]{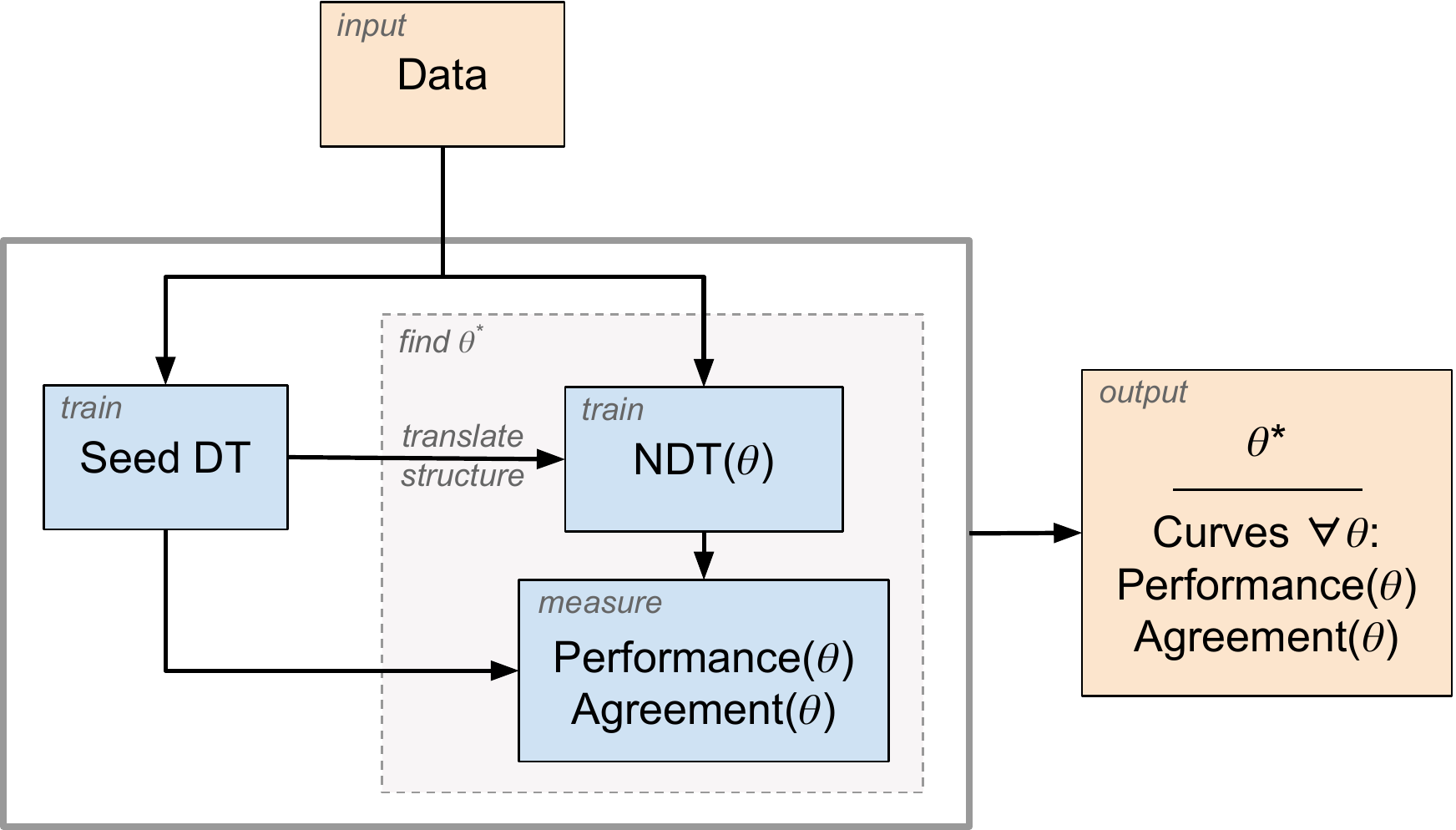}
\vspace{-0.5em}
\caption{Outline of the proposed method.}
\vspace{-1.5em}
\label{fig:outline}
\end{figure}

During the exploratory phase of data analysis, and before choosing a model to fit it to the data, it is interesting to know whether the data can be sufficiently summarized with decision boundaries composed of a set of `hard' or `rigid' rules, which can be interpreted by humans. Rejecting this assumption would mean that it is necessary to have a certain degree of smoothness to the decision boundaries in order to capture better the structure of the dataset.
Typically, one would simply compare members of different model families and select one through a procedure such as cross-validation (CV). In machine learning, where it is more usual to focus on the prediction capacity of models, CV is a widespread procedure both for model and algorithm selection \cite{Ding2018}. Indeed, CV is easy to implement and simplifies the comparison of different models
based on the variability of a chosen performance metric.

From a higher level point-of-view, automated machine learning (auto-ML) or meta-learning methods can be used to first decide which type of  algorithm can be suitable for a dataset, and then to train the final model on it. Indeed, packages such as Auto-WEKA \cite{Kotthoff2019} and Auto-sklearn \cite{Feurer2015} provide tools to automate the algorithm and model selection process. They apply Bayesian optimization and meta-learning procedures \cite{Feurer2015,Reif2012} to select the most appropriate algorithm and its parametrization according to a metric, and within a user-defined budget. Other methods aim to use data characterizations to obtain insights into what kind of data mining algorithm is suitable for a dataset. These characterizations can be statistical and information-theoretical measures to be employed as input, and then the aim is to learn their association with the algorithms' performance for the data. Users can interpret these methods, \eg by producing decision rules using the C5.0 algorithm \cite{Ali2006}, or a self-organizing map to cluster various datasets according to their characteristics \cite{Smith2002}. Other, more complex, data characterizations (or meta-features) have been proposed to describe the problem at hand \cite{Lindner1999,Pimentel2019}.
Closer to our work, the approach of \cite{Peng2002} extracts meta-features from a dataset's inducted decision tree, which attempt to capture learning complexity of the dataset.
These methods, however, require a database of use-cases (datasets and their associated preferred algorithms) whose clustering would provide general guidance on model selection, and might need to be retrained when adding new use-cases.

The aforementioned existing methods aim mostly at deciding among candidate models, and eventually train a good performing final model. In this sense, they do not provide the user with direct insights regarding the complexity of the underlying structure of the data itself, which is something generally less studied in the literature, and it is exactly the main point of focus of this work. Specifically, we propose an exploratory procedure to help the user assessing the expressive power needed for producing efficient classification boundaries for a given dataset. This procedure is meant to be followed to better understand the dataset, prior to selecting the set of models to be further explored through model selection techniques. The procedure can be directly applied to a dataset and does not require prior knowledge for the input data or processing of external data.

More specifically, our idea for assessing the expressive power needed for a dataset is to challenge the decision boundaries produced by a rigid trained model. This is achieved by relaxing progressively its decision boundaries, and evaluating in a controlled way how flexible these need to become so that they fit better to the data. To realize this idea, we use a typical Decision Tree (DT) for the initial decision boundaries, as it is a simple, interpretable, and a naturally rigid model.
The proposed procedure is outlined in \Fig{fig:outline}: it starts with the decision boundaries produced by a reference DT trained on the input dataset, also called seed DT. The seed DT initializes a Neural Decision Tree (NDT) \cite{Biau_2018}, which inherits the DT's decision boundaries. By its definition, an NDT is a special type of Neural Network that can be initialized by a DT, where the smoothness of the activation functions can be controlled. What we put forward is the idea that, by training an NDT, it becomes possible to measure two things: its `departure' from the seed DT in terms of disagreement at the decision level, and the evolution of any performance metric as a function of the allowed smoothness of the decision boundaries.
We show with experiments on real and synthetic data that the indicators provided by our data exploration procedure are meaningful for the classification task, and we illustrate with examples how users can interpret them in practice.

\section{Background}\label{sec:background}

In this section, we present the tools we use to build our procedure.
First, the core of the proposed method is the gradual relaxation of the decision boundaries produced by a rigid model. The algorithm to be used to perform this relaxation needs to offer the possibility of controlling the expressive power of the final model by tuning a small number of parameters. We propose this to be done using a Neural Decision Tree \cite{Biau_2018}, which will be presented in Subsec.\,A. Once a model with more flexible decision boundaries is obtained, we can measure its `departure' from the initial rigid one. Subsec.\,B describes metrics to evaluate the difference between two models.

\inlinetitle{A. Neural Decision Trees (NDTs)}{.}
An NDT is a neural network (NN) whose architecture and weights initialization are obtained directly from an input DT \cite{Biau_2018,Balestriero2017,Lu2020},
which we call here `seed DT'. The NDT variant we use is the one from \cite{Biau_2018}, which we extend to classification tasks. The hyperparameters of this NDT type allow us to control the smoothness of its activation functions, which in fact is a proxy for controlling the smoothness of the decision boundaries.

An important NDT feature is that there is no need for the user to search for the right network architecture for each dataset, \ie the  number of layers or the number of neural units in each layer. Its generally shallow architecture may be too restrictive for complex problems, but this feature is seen as an advantage for our purpose.
An NDT is always formed by four layers: an input layer, two hidden layers, and an output layer. The connections between the layers encode the information extracted from the seed DT. For a dataset with $d$ features and a seed DT with $K$ leaves, we get the following architecture and weights initialization:

\vspace{2em}
$\mySqBullet$~\emph{Input layer.} As usual, it has $d$ neurons corresponding to the data features.

$\mySqBullet$~\emph{First hidden layer}:
It consists of $K-1$ neurons, each one representing a split node of the seed DT. A split condition of a node refers to a feature and a threshold on its value. The NDT encodes this information in the weight and bias matrices of the connections between this layer and the input layer.

$\mySqBullet$~\emph{Second hidden layer}:
It consists of $K$ neurons, one for each leaf of the DT. Then, the connections between the neurons of this layer and those of the previous layer encode the positions of the leaves with respect to each split node. The elements of the weight and bias matrices encode the root-to-leaves paths present in the DT structure.

$\mySqBullet$~\emph{Output layer}:
For a classification task, this layer contains the observed probability of an instance belonging to a class, according to its leaf membership.

$\mySqBullet$~\emph{Activation functions}:
In a DT, the splits, as well as the instance-to-leaf memberships, are crisp. For an NDT to behave like a DT, its activation functions have to be crisp as well. However, a crisp function is not differentiable, and hence it would not be possible to train the NDT using backpropagation. To mitigate this problem, it is proposed to approximate each crisp threshold function of the trees with the function (see \cite{Biau_2018}):

$  \sigma: \ x \mapsto \tanh(\gamma x)$.

 The parameter $\gamma$ allows controlling the smoothness of the $\sigma$ function: the higher the value of $\gamma$, the steeper the curve of $\sigma$ gets. Moreover, the same $\sigma$ form is used as an activation function for both NDT hidden layers, but with a different value of the $\gamma$ parameter in each case, which we denote by $\gamma_1$ and $\gamma_2$ that have the index of the respective hidden layer.

\begin{figure}[t]\small
  \centering
\begin{minipage}{\columnwidth}\centering
\begin{subfigure}{.23\textwidth} \centering
    \includegraphics[width=0.9\columnwidth, trim=0 17cm 11.7cm 1cm, clip]{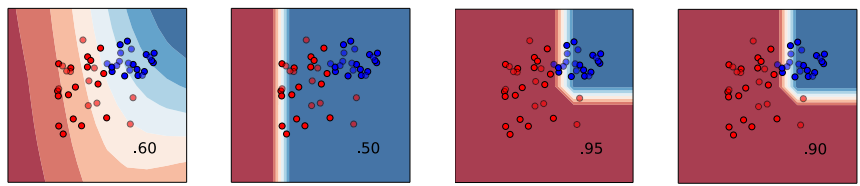}
		\caption{\scriptsize$\gamma_1=1$; $\gamma_2=1$}
    \label{fig:ndt_classif_a}
\end{subfigure}%
\hspace{3mm}%
\begin{subfigure}{.23\textwidth} \centering
    \includegraphics[width=0.9\columnwidth, trim=11.7cm 17cm 0 1cm, clip]{images/NDT_classif.pdf}
		\caption{\scriptsize$\gamma_1=1$; $\gamma_2=100$}
    \label{fig:ndt_classif_b}
\end{subfigure}%
\hspace{1.0mm}%
\begin{subfigure}{.23\textwidth} \centering
    \includegraphics[width=0.845\columnwidth, trim=0.57cm 6.3cm 11.7cm 12cm, clip]{images/NDT_classif.pdf}
		\caption{\scriptsize$\gamma_1=100$; $\gamma_2=1$}
    \label{fig:ndt_classif_c}
\end{subfigure}%
\hspace{2.3mm}%
\begin{subfigure}{.23\textwidth} \centering
    \includegraphics[width=0.9\columnwidth, trim=11.7cm 6.35cm 0 12cm, clip]{images/NDT_classif.pdf}
		\caption{\scriptsize$\gamma_1=100$; $ \gamma_2=100$}
    \label{fig:ndt_classif_d}
\end{subfigure}%
\end{minipage}
    \vspace{-3mm}
    \caption{Decision boundaries for data classification, obtained with NDTs that are all initialized with the same seed DT, whereas with different parameter values $\gamma_1$ and $\gamma_2$ that influence the smoothness and precision of the boundaries of the trained models. For parameter values, the NDT decision boundaries coincide with those of its seed DT.}
    \label{fig:classif}
    \vspace{-1em}
\end{figure}

As depicted in \Fig{fig:classif}, these parameters adjust the flexibility of the trained NDT model, which concerns its ability to learn smoother decision boundaries. Starting from a single DT, several models of variable flexibility can be generated by fixing the seed DT and then training NDTs that differ only in their $\gamma_1$ and $\gamma_2$ parameter values.

\inlinetitle{B. Metrics}{.}
One way to compare two classification models is with respect to their prediction capacity. This can be measured in various ways according to the kind of classification error that is minimized. We denote this performance measure by $\performance$. Nevertheless, two models can have the same performance on a dataset, even though they may use a decision boundary of different structure.
Measuring the distance between two models based on their decision structure is a challenging problem, especially when comparing models from different families, since that would require a meaningful representation for distance calculation.
On the other hand, comparing the behavior of two models can be model-agnostic, that is when one quantifies how much the models agree
or disagree on their predictions on the same dataset.
For classification tasks, in particular, several metrics have been introduced for pairwise model comparison \cite{Zhou_2012}.
We specifically measure the decision agreement, denoted by $\agreement$, between two models as a proxy to measure what we term in this work as the `departure' of one model from the other. Finally, the average value of the metrics over a number of experiments are denoted by $\overbar{\agreement}$ and $\overbar{\performance}$.

\section{Proposed method}\label{sec:method}

\inlinetitle{A. Outline}{.}
The main idea of the proposed data exploration procedure is to start from a DT whose training hyperparameters have been tuned (\eg using CV). This is taken as a seed DT, and its structure is transferred to an NDT with a parameter set $\theta$ (see \Fig{fig:outline}), including the batch size, the number of training epochs, type of optimizer, and the $\gamma_1$ and $\gamma_2$ values. During the procedure, the latter two are the only hyperparameters of the NDT that are not fixed. Therefore, finding $\theta^*$ in our case means finding the values of $\gamma_1$ and $\gamma_2$ for which the best average performance is observed.

$\gamma_1$ and $\gamma_2$ control the smoothness of the NDT activation functions: for higher values (\eg 100) the NDT behaves very similarly to the seed DT, while for lower values it gets closer to an NN with hyperbolic tangent activation functions (see \Fig{fig:ndt_classif_c}). Varying progressively $\gamma_1$ and $\gamma_2$ from higher to lower values \emph{causes a progressive departure of the NDT model}: from the initial DT to an altered model that is relaxed and closer to an NN model. It is then possible to detect the point where better performance is obtained with respect to $\performance$, and also to measure using the metric $\agreement$ how far from a DT is the best NDT model obtained with $\theta^*$, which we also denote by $NDT^* := NDT(\theta^*)$. Note that while we can measure the departure from a specific seed DT, we cannot (as of yet) conclude that the NDT is departing from all the DTs that could be induced from the dataset. Hence the need of repeating the procedure with different seed DTs and evaluating the average behavior of DTs and NDTs, as well as the differences in their distributions, with statistical tests.
The proposed method searches for $\theta^*$ and its subsequent interpretation, as well as that of $\overbar{\performance}(NDT^*)$ and $\overbar{\agreement}(NDT^*)$. This is facilitated by graphical plots depicting the evolution of the metrics as a function of the values of the NDT parameters, over $n$ repetitions.

\inlinetitle{B. Relationship between $\gamma_1$ and $\gamma_2$}{.}
As $\gamma_1$ has a bigger impact on the global shape of the decision boundaries, and in order to simplify the procedure and provide easy-to-interpret results, we link the value of $\gamma_2$ to that of $\gamma_1$ such that $\gamma_2 = f(\gamma_1)$.
In the work \cite{Biau_2018} that originally presented the type of NDT we use, the general rule suggested is to use $\gamma_1 \gg \gamma_2$, since a smoother activation function in the second hidden layer allows for stronger weight corrections in the first hidden layer during backpropagation. An example of how having $\gamma_1 < \gamma_2$ can cause the loss of the initial information obtained from the seed DT and prevent the correct training of an NDT, can be seen in \Fig{fig:ndt_classif_b}.

Our particular goal is to start with an NDT that coincides with its seed DT, and then progressively relax the produced NDT's decision boundaries through training. This translates to $\gamma_2$ being of the same order of magnitude as $\gamma_1$. This is because if a much lower value of $\gamma_2$ is used, it would mean that, while the parts of the NDT corresponding to the information of the seed DT's inner nodes (first NDT layer) would progressively be adjusted, the paths between them would be able to be updated in an unconstrained way. However, this is not desired.

Note that there is no known optimal function relating $\gamma_1$ and $\gamma_2$. To present our proof of concept, we decided to link the hyperparameters with the function $f:x\mapsto \sqrt[q]{x}$, which allows us to respect the conditions stated previously.
In what follows, we use $\gamma$ to actually refer to $\gamma_1$, and then $\gamma_2$ is computed internally by:%

\begin{equation}\label{eq:gammas}
  \gamma_2 = f(\gamma_1) = \sqrt[q]{\gamma_1}, \text{  with  } q = 1.1\ .
\end{equation}

\inlinetitle{C. The algorithm of the procedure}{.} \Eq{eq:gammas} reduces the process of finding $\theta^*$ to that of estimating the optimal $\gamma^*$. As we are interested in a progressive departure from any given DT, we can produce multiple seed DTs, and for each one of those to build several NDTs by decreasing gradually the value of $\gamma$. The $36$ tested values are in the ordered set $\Gamma =  \{9000, 8000, \mydots, 1000\} \cup \{900, 800, \mydots, 100\} \cup \{90, 80, \mydots, 10\} \cup \{9, 8, \mydots, 1\}$. %
For each trained NDT, we measure the chosen performance metric $\performance$ and the agreement $\agreement$ (see \Sec{sec:background} -- B). The overall procedure is detailed in \Alg{alg:method}.

\begin{algorithm}[t]
\scriptsize
\SetAlgoLined
  \KwIn{a dataset $D$, a set $\Gamma$ of values to test for $\gamma$}
\KwOut{the optimal $\gamma^*$ value of $\gamma$, the mean agreement $\overbar{\agreement}(NDT, DT)$, and the and mean respective performance $\overbar{\performance}(NDT)$ and $\overbar{\performance}(DT)$}
\smallskip
\smallskip
  find the hyperparameters $h$ of a DT model for the dataset

  \For{$i=1,...n$ \emph{times}}{
    form the $D_{\text{train}}$, $D_{\text{test}}$, and $D_{\text{valid}}$ sets from $D$

    build the new $DT_i := DT_i(h)$ with $D_{\text{train}}$

    measure $\performance(DT_i)$ on $D_{\text{test}}$

    \ForEach{$\gamma \in \Gamma$}{

      initialize $NDT_{i,\gamma} := NDT(\gamma)$ using $DT_i$

      fit $NDT_{i,\gamma}$ to $D_{\text{train}}$ and regularize it with $D_{\text{valid}}$

      measure $\agreement(NDT_{i,\gamma}, DT_i), \performance(NDT_{i,\gamma})$ on $D_{\text{test}}$

    }
  }

  $\overbar{M_{DT}} = \frac{1}{n}\sum_{i=1}^n \performance(DT_i)$

  $\overbar{M} = \left[\frac{1}{n}\sum_{i=1}^n \performance(NDT_{i, \gamma})\right]_{\gamma \in \Gamma}$

  $\overbar{A} = \left[\frac{1}{n}\sum_{i=1}^n \agreement(NDT_{i, \gamma}, DT_i)\right]_{\gamma \in \Gamma}$

  $\gamma^* =  \argmax_{\gamma \in \Gamma}\  \overbar{M}_\gamma$

  \Return{$\gamma^*$, $\overbar{A}$, $\overbar{M}$, $\overbar{M_{DT}}$}

  \caption{Pseudocode of the proposed method}
  \label{alg:method}
  \end{algorithm}

 As we mainly aim to show a proof of our concept, we use a simple way to optimize $\gamma$: by testing $n$ times the performance of $NDT(\gamma)$ for $\gamma\in\Gamma$, and then by selecting the value that provides the best average performance. \Alg{alg:method} can be computationally expensive for a large $\Gamma$ set, as that would lead to the training of $n\times|\Gamma|$ NNs. However, by using NDTs and by relating $\gamma_2$ to $\gamma_1$ (\Eq{eq:gammas}), there is no need to also search for the optimal NN architecture, and hence we reduce the parameter space to be searched. Note also that the initialization of the NDT weights by a seed DTs provides a warm start, which makes the NDT training to be less sensitive to randomness and generally to converge faster compared to usual initialization techniques (e.g. random uniform weights).

\inlinetitle{D. Using comparative metrics between models}{.}
Given a trained NDT that was initialized by a specific seed DT, we measure its departure with respect to its seed DT as the decision agreement $\agreement$ between the two models.

More specifically, the agreement metric used for the considered  classification task is Cohen's $\kappa$ statistic \cite{Cohen1960}, which calculates the extent to which the class predicted by two classifiers agree while also taking into account the probability for them to agree by chance:
  $\kappa = \frac{p_o-p_e}{1-p_e}$, where $p_o$ is the probability that the two classifiers do agree, and $p_e$ the probability that they agree by chance. When the models agree completely, $\kappa = 1$; when they agree by chance, $\kappa = 0$; if their agreement is less than what is expected by chance, then $\kappa < 0$.
By abusing the notation, we write as $\agreement(NDT(\gamma))$ the agreement evaluation for an NDT and its seed DT. %
Finally, the performance metric $\performance$ that is used to measure and compare the performance of different models is their classification accuracy.

\section{Experiments}\label{sec:exps}

\inlinetitle{Datasets}{.}
The proposed method is tested on $8$ datasets, $1$ synthetic and $7$ containing real data taken from the UCI repository. The characteristics of these datasets are summarized in \Tab{tab:datasets_1}. Some of them  were chosen because either rule-based models (such as DTs) or more expressive models (such as NNs) are better suited for them. First, the synthetic \dset{sim\_1000\_3} dataset contains two non-linearly separable classes, each of them generated by a Gaussian distribution. Regarding real data, the Gastrointestinal Lesions in Regular Colonoscopy dataset (\dset{lesions}) was chosen as it is high-dimensional and in such cases DTs can be better in selecting only the most informative features \cite{Brown_1993}.
The \dset{mushroom} dataset is interesting as it can be accurately modeled using simple rules. The rest of the datasets used are cases where the underlying structure is not as clear as above: the Spambase (\dset{spam}), the Congressional Voting Records (\dset{votes}), Student Performance (\dset{student-math}), and the Wine (\dset{wine}) datasets, as well as a part of the Vehicle Silhouettes dataset (\dset{xab}).
The last three datasets were transformed into binary classification problems. The implementation of the compared methods, and the datasets used, are available online\footnote{See: \url{https://kalogeratos.com/material/DTvsNDT}.}.
\begin{table*}[h]
  \begin{scriptsize}
  \begin{minipage}{\textwidth}
      \centering
      \caption{Datasets used to obtain the experimental results.}
      \label{tab:datasets_1}
      \begin{tabular}{l||r|r|r}
        \toprule
        \textbf{Dataset}  &  \textbf{\#Feats} &
        \textbf{\#Inst.} & \textbf{DT depth}  \\
        \midrule
         \dset{sim\_1000\_3} & 3 & 1000 & 4 \\
         \dset{wine} & 13 & 178 & 3 \\
         \dset{votes} & 16 & 435 & 2 \\
         \dset{xab} & 18 & 94 & 6  \\
         \dset{mushroom} & 23 & 8124 & 5 \\
         \dset{student-math} & 31 & 395 & 3 \\
         \dset{spam} & 58 & 4601 & 5 \\
         \dset{lesions} & 371 & 152 & 4 \\
        \bottomrule
      \end{tabular}
  \end{minipage}
  \end{scriptsize}
	\vspace{-1.5em}
  \end{table*}

\inlinetitle{Experimental pipeline}{.}
For each dataset, we first used CV to determine the depth of the DTs to be used as seeds (reported in \Tab{tab:datasets}), which was found to be the hyperparameter with the biggest impact on DT performance. Aside from removing the categorical variables (except for \dset{mushroom}, where all its variables are categorical and were encoded, similarly to \dset{student-math}) and the instances with missing data, no other preprocessing preceded.

For the NDTs, for all datasets, we set the number of training epochs to $100$ and the batch size to $32$. We use the Adam optimizer with default parameters, and for regularization we use early stopping with a patience of $10$ epochs.

To generate the training, validation, and test sets we use repeated random subsampling with proportions 50\%\,/\,25\%\,/\,25\% respectively, which we repeat $30$ times (\ie $n = 30$). We use stratification to ensure that classes are in the same proportions in the sets.

We additionally compare the results obtained with the performance of an NN architecture chosen via a typical CV procedure. The parameters we investigated are: the depth ($D = \{1, 2, 3\}$), the width for all layers ($W = \{2, 3, 4, 5, 6, 7\}$), and the activation function to be used for all hidden layers ($\tanh$ or ReLU). This corresponds to training $36$ NNs for each dataset, hence the computational budget is comparable to that used by our method when $|\Gamma| = 36$.

Finally, we evaluate the significance of the differences of the observed performance using statistical tests. We use Wilcoxon's signed rank test \cite{Wilcoxon1945} (to which we refer to as Wilcoxon's test in the following sections), since the paired samples we compared can not be assumed to be normally distributed (as per Shapiro-Wilk's test). The null hypothesis we test is: the median of the population of differences between the paired data to be zero ($H_0$), and the two-sided alternative hypothesis is: the median of the population of differences to be different from zero ($H_a$). These hypotheses are tested with a risk $\alpha = 0.05$.

\inlinetitle{Results}{.}
The method outputs $\gamma^*$, $\overbar{\performance}(NDT^*)$, and $\overbar{\agreement}(NDT^*)$. These values are then to be interpreted so that we get insights whether the dataset would need more flexible decision boundaries than those of a hard set of rules. We present here detailed examples of this interpretation for three datasets (\dset{sim\_1000\_3}, \dset{lesions} and \dset{spam}). We also present the results of tests that measure the statistical significance of the obtained results, and how they compare to a CV done in a limited set of NNs.

\begin{table}[t]
\caption{Summary of the experimental results. The results are the average of $30$ iterations, and `Imp.' indicates whether an NDT achieved (on average) an improvement over the performance of its seed DT.}
\label{tab:datasets}
\begin{scriptsize}
\begin{minipage}{\textwidth}
\centering
\begin{tabular}{l||r|r|r|r|c|r|r}
\toprule
 \textbf{Dataset}  & \bm{$\gamma^*$}  & \bm{$\overbar{\performance}(DT)$}
        & \bm{$\overbar{\performance}(NDT(\gamma^*))$}
        & \bm{$\overbar{\performance}$} \textbf{diff.}
        & \textbf{Imp.} & \bm{$\overbar{\agreement}(NDT(\gamma^*))$}   & \bm{$\overbar{\performance}(NN)$}\\
\midrule
sim\_1000\_3 & 3 & 0.906 (0.008) & 0.989 (0.008) & -0.083 (0.010) & \checkmark & 0.821 (0.020) & 0.990 (0.009) \\
wine & 8000 & 0.962 (0.017) & 0.956 (0.024) & 0.007 (0.016) &        & 0.976 (0.034) & 0.627 (0.076) \\
votes & 50 & 0.977 (0.008) & 0.971 (0.020) & 0.006 (0.019) &        & 0.974 (0.060) & 0.748 (0.138) \\
xab & 60 & 0.871 (0.041) & 0.865 (0.047) & 0.006 (0.018) &        & 0.982 (0.035) & 0.532 (0.066) \\
mushroom & 4 & 0.983 (0.002) & 1.000 (0.000) & -0.016 (0.002) & \checkmark & 0.967 (0.003) & 0.969 (0.015) \\
student-math & 5 & 0.735 (0.014) & 0.721 (0.022) & 0.014 (0.027) &        & 0.601 (0.266) & 0.675 (0.010) \\
spam & 1 & 0.916 (0.004) & 0.941 (0.006) & -0.025 (0.006) & \checkmark & 0.845 (0.021) & 0.920 (0.059) \\
lesions & 8000 & 0.891 (0.025) & 0.889 (0.037) & 0.003 (0.028) &        & 0.892 (0.127) & 0.719 (0.111) \\
\bottomrule
\end{tabular}
\end{minipage}
\end{scriptsize}
\vspace{-1.5em}
\end{table}

 \begin{table*}[h]
\scriptsize
\begin{minipage}{\textwidth}
\centering
\caption{Results from statistical Wilcoxon's signed rank tests comparing pairwise the accuracies of the models NDTs, DTs, and NNs.}
\label{tab:tests}
\begin{tabular}{lrr|c|r|c|r|c}
\cmidrule[\heavyrulewidth]{3-8}
& & \multicolumn{2}{c}{\textbf{NDTs \vs DTs}} & \multicolumn{2}{|c}{\textbf{NDTs \vs NNs}} & \multicolumn{2}{|c}{\textbf{NNs \vs DTs}} \\
\midrule
\textbf{Dataset} &\multicolumn{1}{||r|}{ \bm{$\gamma^*$}} & \textbf{$p$-value} & \textbf{Reject} $\mathbf{H_0}$ & \textbf{$p$-value} & \textbf{Reject} $\mathbf{H_0}$ & \textbf{$p$-value} & \textbf{Reject} $\mathbf{H_0}$ \\
\midrule
\dset{sim\_1000\_3} & \multicolumn{1}{||r|}{3} & 1.65e-06 & \checkmark & 5.49e-01 &         & 1.64e-06 & \checkmark\\
\dset{wine} & \multicolumn{1}{||r|}{8000} & 8.38e-02 &         & 1.63e-06 & \checkmark & 1.55e-06 & \checkmark\\
\dset{votes} & \multicolumn{1}{||r|}{50} & 4.89e-01 &         & 2.29e-06 & \checkmark & 1.66e-06 & \checkmark\\
\dset{xab} & \multicolumn{1}{||r|}{60} & 2.15e-01 &         & 1.65e-06 & \checkmark & 1.62e-06 & \checkmark\\
\dset{mushroom} & \multicolumn{1}{||r|}{4} & 1.56e-06 & \checkmark & 1.73e-06 & \checkmark & 2.36e-05 & \checkmark\\
\dset{student-math} & \multicolumn{1}{||r|}{5} & 1.81e-02 & \checkmark & 1.89e-06 & \checkmark & 1.61e-06 & \checkmark\\
\dset{spam} & \multicolumn{1}{||r|}{1} & 1.72e-06 & \checkmark & 2.70e-05 & \checkmark & 5.53e-05 & \checkmark\\
\dset{lesions} & \multicolumn{1}{||r|}{8000} & 1.49e-01 &         & 1.61e-06 & \checkmark & 1.66e-06 & \checkmark\\
\bottomrule
\end{tabular}
\end{minipage}
\vspace{-1.5em}
\end{table*}

\begin{figure}[t]\centering
\small
\hspace{-13em}
\begin{minipage}{\columnwidth}\centering
\begin{subfigure}{.45\textwidth} \centering
    \includegraphics[width=0.95\columnwidth, viewport=10 10 350 298,clip]{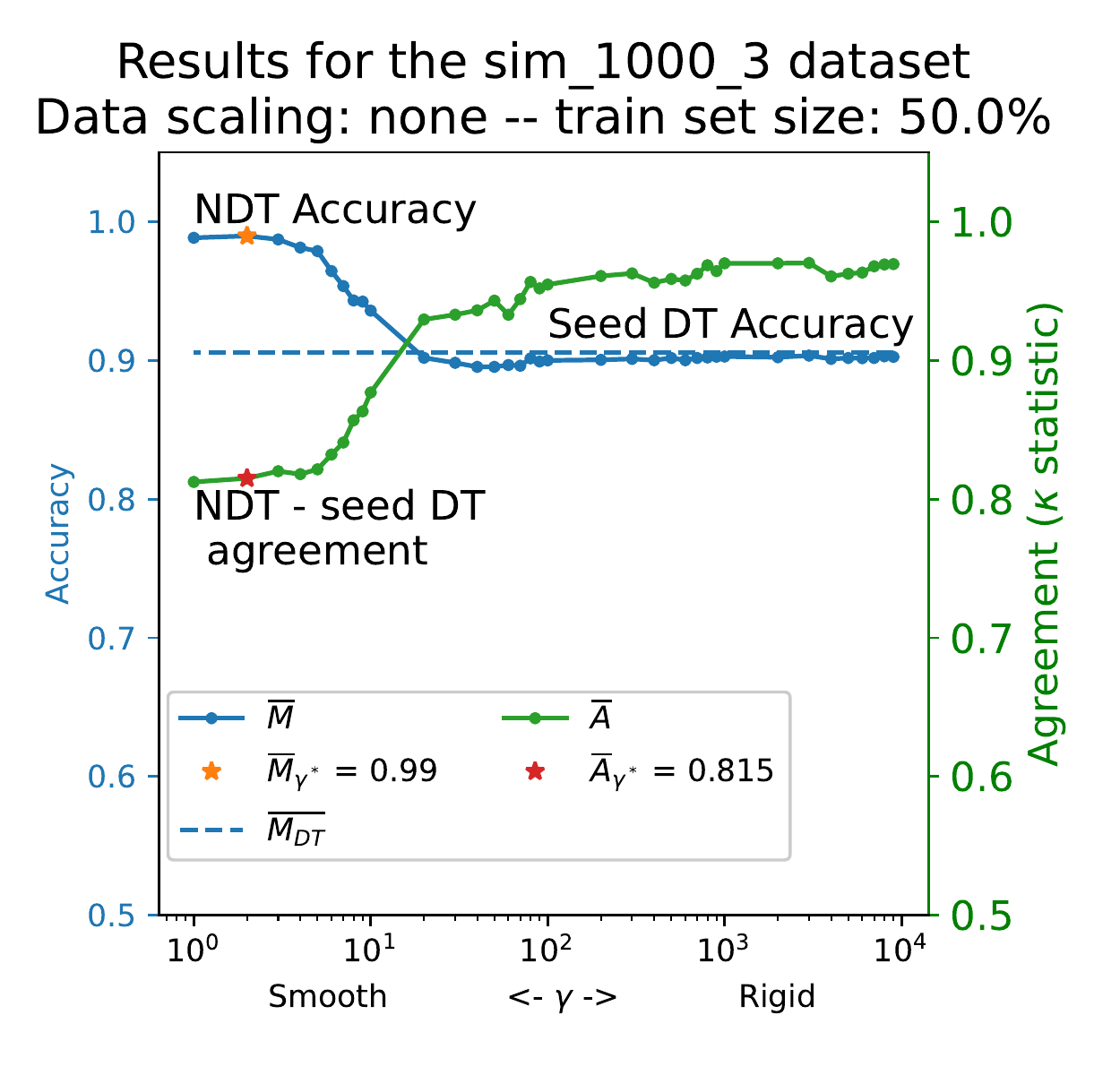}
		\caption{}
    \label{fig:results}
\end{subfigure}%
\begin{subfigure}{.45\textwidth} \centering
    \includegraphics[width=0.95\columnwidth, viewport=10 10 350 298,clip]{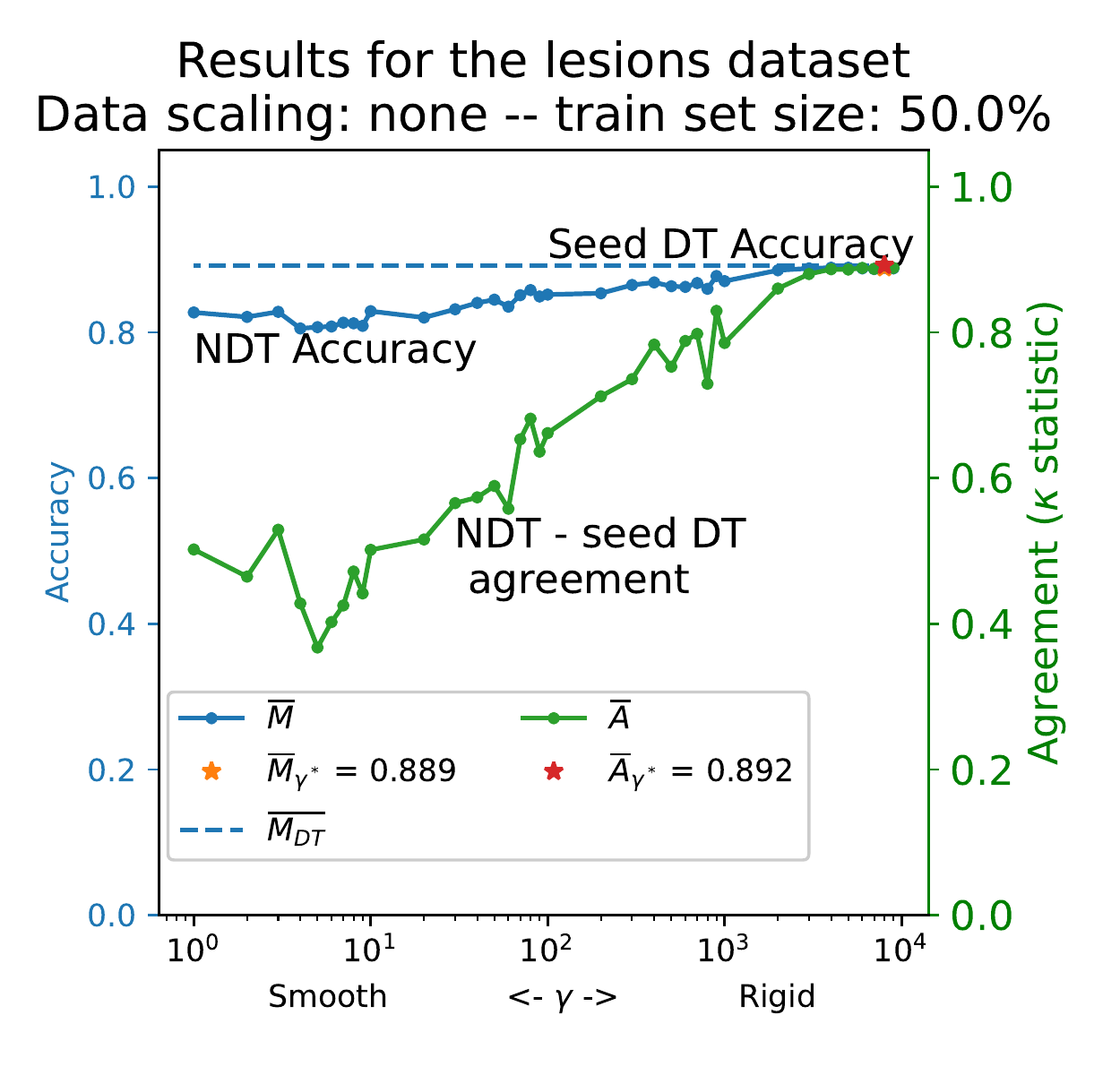}
		\caption{}
    \label{fig:results_lesions}
\end{subfigure}%
\begin{subfigure}{.45\textwidth} \centering
    \includegraphics[width=0.95\columnwidth, viewport=10 10 350 298,clip]{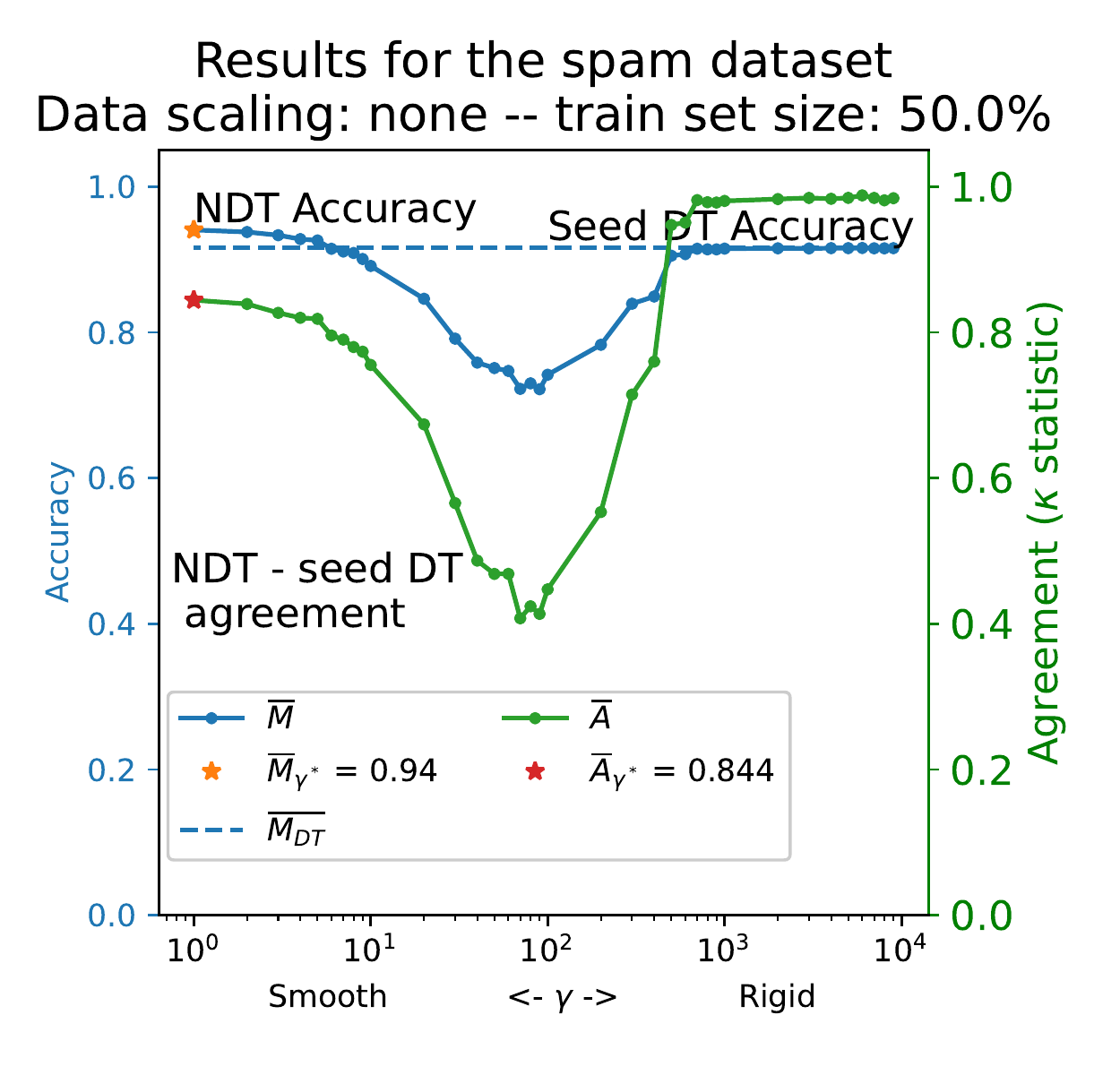}
    \caption{}
		\label{fig:results_spam}
\end{subfigure}
\end{minipage}
\vspace{-2mm}
    \caption{Progressive relaxation (right-to-left along x-axis, as $\gamma$ reduces) of the NDT decision boundaries with respect to its seed DT.
    No data scaling is used; the training set is 50\% of each dataset. \textbf{(a)} \dset{sim\_1000\_3} dataset -- As the boundaries are getting relaxed, the NDT performs better, but agrees less with the seed DT, suggesting that a more flexible model might be more adapted. \textbf{(b)} \dset{lesions} dataset -- Both the agreement and the accuracy of the NDT decrease with the value of $\gamma$, meaning that a relaxation of the initial DT's boundaries is not beneficial. \textbf{(c)} \dset{spam} dataset -- As the boundaries get relaxed, the NDT performs better, and also agrees more with the seed DT, which means that the relaxation is beneficial when the rigid boundaries ($\gamma$ is high) are sensitive to small changes to which it is hard for them to adapt through training.}
\vspace{-1.5em}
\end{figure}

\inlinetitle{Examples}{.}
\Fig{fig:results} shows the average performance of the NDT with respect to $\gamma$, and compares it to that of the initial seed DT. It tells us that, as expected, for higher values of $\gamma$ the NDT has an average performance that is close to that of its seed DT, and as $\gamma$ decreases the performance of the NDT varies. In this case, the accuracy increases monotonically and reaches its maximum for $\gamma = 3$. For this value, $\sigma$ is closer to a $\tanh$ function and thus the NDT decision boundaries are smooth. This is a case where the specific needs of the user will determine which should be the next steps to take. Indeed, the high agreement between $NDT(\gamma^*)$ and the DT ($0.821$, see \Tab{tab:datasets}) indicates that a DT and a more flexible model may not differ much decision-wise, so the latter can be more promising to explore if one is interested in obtaining better performance at the cost of sacrificing the explainability. Nevertheless, if one can afford to sacrifice a bit in terms of model accuracy to gain understanding, the former can suffice.

\Fig{fig:results_lesions} shows both the performance and the agreement decreasing as the value of $\gamma$ decreases. Here, the best average accuracy for the NDT is reached when $\gamma^*=8000$. In this case, the function $\sigma$ is practically a threshold function, and hence the NDT will behave similarly to a DT. In contrast, there is no improvement over the seed DT, even for $NDT^*$.
Our interpretation of these results is that the dataset can be described by a model with rigid decision boundaries, such as a set of decision rules or a DT.

Similar reasoning can be used for the rest of the datasets, based on the results of \Tab{tab:datasets}. Similar graphs can be drawn for all the datasets, however, their interpretation might be less straightforward than for the examples of \Fig{fig:results}.
The data preprocessing can also greatly affect the shape of the plots.

\Fig{fig:results_spam} is an example where the interpretation is more complex. In this figure, the agreement and the classification performance do not have a monotonic evolution, but their curves follow similar trends. This can be explained by the fact that the agreement measure we use evaluates the proportion of the instances where an NDT and its seed DT make the same decisions, and this is influenced by the quality of the models being compared. In \Fig{fig:results_spam}, we can see that the seed DTs have a good average accuracy, and so when the derived NDTs are more accurate, they are bound to make similar decisions, and vice-versa. When the NDTs are less accurate, it is natural that these models make different predictions for a significant part of the dataset.
When $\gamma$ is lower ($ < 100$), we observe an increase in the accuracy of the NDTs, with the best accuracy being reached when $\gamma^* = 1$. In this case, the information we have at hand indicates that the dataset can be modeled using a DT, but its accuracy is sensitive to changes in its structure. When $\gamma$ is low enough, it is easier for the NDT to adapt to these changes, since its decision boundaries get sufficiently flexible. It can thus progressively approach a better local minimum of the loss function. In conclusion, a model with a higher expressive power is worth exploring in this case, although the agreement with a DT might be high (it is $0.845$ as shown in \Tab{tab:datasets}).

\inlinetitle{Testing for statistical significance}{.}
In support of our experimental results, we performed statistical tests to confirm the significance of the difference in the median performance of the seed DTs and the $NDT(\gamma^*)$ for each dataset.
We also present tests to show the significance of the difference between DTs and NNs, and between NNs and $NDT(\gamma^*)$.

\Tab{tab:tests} reports the results of Wilcoxon's tests, which show that when $\gamma^*$ is low, the hypothesis $H_0$ that the median difference in the performance of the two models that are compared is 0. $H_0$ is rejected when the $p$-value is lower than $0.05$. %
We observe that we can reject $H_0$ and accept $H_a$ for half of the datasets. These are the datasets where $\gamma^*$ is the lowest, meaning that even though NDTs are initialized by the DTs, a statistically significant gain (or loss) is achieved by the departure of the NDT from the DT. The cases where we cannot reject $H_0$ are those where $\gamma^*$ is higher, and $\agreement$ is very high, as shown in \Tab{tab:datasets}. The need for departure from a DT structure by relaxing its decision boundaries is reflected by the $\agreement$ and $\gamma^*$ values.

Furthermore, when NNs and NDTs are compared, our tests show that $H_0$ can be rejected in most cases, except for the \dset{sim\_1000\_3}, where $\gamma^*$ is low and $\agreement$ is relatively low. Notice also that in most cases the performance of the NDT is better than that of the NN found via CV.

Simply comparing the performances of DTs and NNs leads to significantly different performance for all datasets, confirming that indeed the DT and NN model spaces are clearly separated for the datasets we use. However, this comparison, is not in itself very informative for the datasets.

\inlinetitle{Note on the computational complexity}{.}
As here we mainly presented a proof of concept, there are aspects that can be further optimized. One of them is the search strategy for $\gamma^*$. Although performing an exhaustive search makes the curves of $\agreement$ and $\performance$ easier to understand by the user, we have observed that these curves evolve monotonically in certain subranges of $\Gamma$. For example, reading right-to-left \Fig{fig:results_spam}, there is first a plateau when $\gamma\in[700,9000]$, then there is a monotonic decrease of the metric values when $\gamma\in[90, 600]$, and finally there is a monotonic increase when $\gamma < 90$. Interestingly, the monotonicity of the observed behaviors is preserved in the range of the same order of magnitude of $\gamma$. This implies that the general behavior of the metrics over $\Gamma$ could be inferred from fewer points, and hence use the presented method more efficiently.

\section{Conclusion}\label{sec:conclusion}
The data exploration procedure we presented in this work allows us to determine how flexible should be the decision boundaries should be for a given classification task, in comparison to those of a rigid model. We propose a controlled way to investigate this by using NDTs initialized by a seed DT, and then progressively relax the NDTs' decision boundaries through training their weights. Furthermore, the agreement metric provides insights about how far it is necessary to depart from the seed DT to achieve a performance gain (if there is one). We demonstrated that our data exploration procedure is meaningful for analyzing the properties of a given dataset, and how the computed metric can be used as indicators by the user. The analysis of the statistical soundness of the results support our findings.

In the future, we plan to work on the optimization of $\gamma_1$, $\gamma_2$ and the enrichment of the experimental results.
Even more important is to investigate further other concrete uses of our procedure that can be beneficial for practitioners.

\subsubsection{Acknowledgements} This work was funded by the Île-de-France Region and additionally by the IdAML Chair hosted at ENS Paris-Saclay. The authors are thankful to Atos for providing an Atos Edge machine to perform numerical experiments.

\bibliographystyle{splncs04}
\bibliography{refs}

\end{document}